\documentclass[letterpaper, 10 pt, conference]{ieeeconf}  

\usepackage{epsfig}
\usepackage{epstopdf}
\usepackage{graphics}
\usepackage{epstopdf}
\usepackage{upgreek}
\usepackage{amssymb}
\usepackage{amsfonts}
\usepackage{amsmath}
\usepackage{refstyle}
\usepackage{xcolor}
\usepackage{subfigure}
\usepackage{todonotes}
\usepackage{mathtools}
\usepackage{algorithm}
\usepackage{algpseudocode}
\usepackage{cite}
\usepackage{cleveref}
\usepackage{url}
\usepackage{leftidx}
\usepackage[detect-all]{siunitx}
\usepackage{makecell}
\usepackage{multirow}
\usepackage{float}
\usepackage{makecell}
\usepackage{booktabs}
\usepackage{wrapfig}

\IEEEoverridecommandlockouts                              

\title{\LARGE \bf
Center-of-Mass-based Robust Grasp Planning for Unknown Objects Using Tactile-Visual Sensors}

\author{Qian Feng$^{1,3}$, Zhaopeng Chen$^{1*,2}$, Jun Deng$^{1}$, Chunhui Gao$^{1}$,\\Jianwei Zhang$^{2}$, and Alois Knoll$^{3}$ 
\thanks{$^{1}$Agile Robots AG, Gilching Germany. \{{\tt\small qian.feng, jun.deng, chunhui.gao\}@agile-robots.com}
        }%
\thanks{$^{2}$TAMS (Technical Aspects of Multimodal Systems), Department of Informatics, Universit\"{a}t Hamburg}%
\thanks{$^{3}$Chair for Robotics and Embedded Systems, Technische Universit\"{a}t M\"{u}nchen, Germany. {\tt\small knoll,qian.feng @tum.de}
		}%
\thanks{*Corresponding author to provide e-mail: zhaopeng.chen@agile-robots.com}%
}

\DeclarePairedDelimiterX{\norm}[1]{\lVert}{\rVert}{#1}
\begin{document}

\maketitle
\thispagestyle{empty}
\pagestyle{empty}

\begin{abstract}
An unstable grasp pose can lead to slip, thus an unstable grasp pose can be predicted by slip detection. A regrasp is required afterwards to correct the grasp pose in order to finish the task.
In this work, we propose a novel regrasp planner with multi-sensor modules to plan grasp adjustments with the feedback from a slip detector. Then a regrasp planner is trained to estimate the location of center of mass, which helps robots find an optimal grasp pose. The dataset in this work consists of 1\,025 slip experiments and 1\,347 regrasps collected by one pair of tactile sensors, an RGB-D camera and one Franka Emika robot arm equipped with joint force/torque sensors. We show that our algorithm can successfully detect and classify the slip for 5 unknown test objects with an accuracy of 76.88\% and a regrasp planner increases the grasp success rate by 31.0\% compared to the state-of-the-art vision-based grasping algorithm.
\end{abstract}

\section{Introduction}
\label{intro}
Robotic grasping and manipulating of unknown objects bring enormous profit to human society. Leveraging the fast advances in deep learning and computer vision, robotic grasping draws increasing interests from industries, e.g., the Amazon Robotic Challenges~\cite{ARCMIT2018}. However, to obtain a robust performance, vision based methods have some limitations. First, the currently proposed methods either learned from realistic data~\cite{Trial-single2015,ARCMIT2018,STNCNN2018,AnchorGrasp2018,Multi-objectCNN2018} or from synthetic data~\cite{dex4.0,Seggrasp2018,MetaGrasp2019}, only involve object shapes or geometries, ignoring many other aspects such as material properties and object mass. Second, the vision-only method is open-loop without information of contacts with objects, thus the robustness is hard to guarantee. 

In consequence, one sensory modality cannot provide enough context to plan a robust grasp. 
Several novel fusion methods of multi-sensor modules are proposed, to learn to detect object geometries~\cite{VioTacGeometry2018}, to help robotic grasp planning~\cite{Gelsight2017, Gelsight2018, VioTacTrash2019}, to execute manipulation tasks like peg insertion\cite{VioTacInsertion2019}. Here we propose a novel algorithm with fusion of multi-sensor modules for unknown objects. 

Meanwhile, grasping and manipulating unknown objects are essential for service robots in the future to execute challenging tasks. For object manipulation, a stable grasp is prerequisite but grasping objects with uncertainties of surface textures~\cite{TactileMaterial2016,viotac-grasp2016} and center of mass~\cite{dynamicCOM2016,Centerofmass}, are still challenging for robotic systems, such as some common tools, e.g., axes and hammers which combine more than one material in one object, are hard to grasp and manipulate because of uneven mass distributions towards simple robot end effectors for example a two-jaw gripper.

A grasp is unstable if the grasp pose is not force-closure/antipodal~\cite{Lizexiang} or the grasp force is too small to provide enough friction~\cite{dynamicCOM2016}, resulting in insufficient contact or slip. Slip detection has been studied since the late 1980s till now with either analytical methods~\cite{Jens2014,CITEC2012,Zhe2015,IIT2016} or data-driven methods~\cite{HMM2012,Filipe2015,Tsinghua2017,SlipLSTM2018,GelsightSlip2018}. With the feedback of slip detection, regrasp can be planned accordingly by applying more grasp force~\cite{Filipe2018,dynamicCOM2016}, when the grasp pose is stable and the contact situation is unchanged. Or with a grasp stability estimator, regrasp can be chosen by scoring the regrasp poses~\cite{Gelsight2018,Tac-Trans2018}. Regrasp planning using feedback of slip detection has advantages of regulating grasp forces and more robustness by identifying a bad grasp with a slip detector instead of pure information from touch. 
To the best of our knowledge, so far, there is no regrasp planner which is based on the slip detection to plan a new stable grasp pose.

In this paper we propose a novel center-of-mass-based robotic grasping algorithm using tactile-visual sensors to grasp unknown objects. The grasp failures are found by slip detection and then a regrasp is planned to improve grasp stability. The proposed algorithm takes object shapes and object mass into consideration using multi-sensor modules. We close the loop of robotic grasping to make it more robust with better performance. We employ Support Vector Machine (SVM) and Long short-term Memory (LSTM) model to extract the tactile features from tactile sensors to detect slip as well as to plan regrasp, and we collect a dataset of 1039 slips and 1347 regrasps from 19 experimental objects. To the best of our knowledge, this is the first work that utilizes multi-sensor modules to plan regrasp based on slip prediction.


We list the main contributions of our paper as follows:
\begin{enumerate}
  \item A novel center-of-mass-based robust grasp planning to grasp unknown objects.  Then to close the loop of grasping, a novel slip detection method is proposed and a novel regrasp planner is trained to estimate the grasp stability by estimating center of mass.
  \item A multi-modal robotic grasping dataset contains tactile sensors and RGB-D images for daily objects. 
  \item The proposed algorithm is proved to be feasible to grasp daily objects with good performance, even for a low-resolution tactile sensor with only $4\times4$ tactile taxels.
\end{enumerate}

\section{Related Work}
\label{sec:relatedwork}

Recently grasp planning has been extensively researched in the past years in the robotic field~\cite{dex2.0,CNNGrasp2013,STNCNN2018,PointNetGPD2018,Gelsight2017,OnlyTactile2018}.


\textbf{Analytical methods}. These methods require an explicit model of the objects and the robot kinematics. First, robots need to ``know" this object previously. To provide this data, some precomputed databases such as Columbia Grasp Database~\cite{Columbia} of 3D objects have been established, where each object is labeled with grasp quality metrics such as grasp wrench space (GWS) analysis~\cite{Springer}. 
Afterwards the best grasp is chosen out of the precomputed grasp poses with the highest score according to the quality metric in~\cite{dex2.0}.
Analytical method only works well if the real-world system fits the assumption of an analytic model without uncertainties, such as unseen objects, un-structured environment.


\textbf{Empirical methods with vision sensors}. Given sufficient data, empirical methods are more robust against uncertainties during robotic grasping~\cite{Review2018}. Recently empirical methods are able to predict grasp pose using RGB-D cameras by either end-to-end learning methods~\cite{CNNGrasp2013,CNNGrasp2015,ARCMIT2018,STNCNN2018,Multi-objectCNN2018} or learning a grasp pose scoring function~\cite{Graspfunction2016,dex2.0,dex4.0,PointNetGPD2018}. 


\textbf{Empirical methods with tactile sensors}. Tactile sensing in robotic grasping can be used to detect object material~\cite{MaterialTactile2016}, estimate force~\cite{Gelsight2018}, detect slip~\cite{Filipe2015,SlipLSTM2018} and estimate grasp stability. 
Considering the limitations of vision, researchers are also discovering how much a tactile sensor can contribute to the grasping performance in combination with vision. Since accurate modeling of contact physics is complex and parameter-dependent, analytical methods~\cite{Jens2014,America2009,Zhe2015} are not able to generalize to unknown objects. 
The tactile sensor can contribute to the grasping performance in two ways, either using the data collected during closing the robot end effector to estimate grasp stability~\cite{Gelsight2017, unsupervise-grasp-stability2017,OnlyTactile2018} or using the data to detect slip~\cite{Jens2014,Zhe2015,IIT2016,CITEC2016,SlipLSTM2018}.





\textbf{Slip detection}. Analytical slip detection methods detect the derivative of the normal force in~\cite{Jens2014,America2009,Zhe2015}, estimate the ratio between friction and the normal force according to the Coulomb friction model by~\cite{Jens2014,Zhe2015,IIT2016} as well as analyze vibrations features~\cite{Jens2014,Zhe2015,CITEC2012}.

Empirical methods become more popular in research because they can handle uncertainties in the environment and can be generalized to unknown objects.
Machine learning methods are used e.g., Hidden Markov Models (HMM)~\cite{HMM2012}, SVM~\cite{Tsinghua2017} and Random Forest~\cite{Filipe2015}, deeper architecture of network e.g., convolutional network by treating tactile feedback as images~\cite{CITEC2016} and LSTM~\cite{SlipLSTM2018}.
In most literature, the experiment regarding slip detection is conducted to keep either the object or the robot end effector static except~\cite{Tsinghua2017}, which is not the typical scenarios for robot grasping application.




Meanwhile, all methods mentioned above to predict grasp stability are still open-loop. Without a regrasp planner, the grasping process has to iteratively "touch" objects until a stable enough grasp pose is found.

\textbf{Regrasp}. One idea of tactile based regrasping is to directly apply more force when the slip is detected only under the assumption that the grasp pose is correct. There are some model-based methods to adjust the grasping force considering the feedback of tactile sensing~\cite{Erik2013,dynamicCOM2016, Filipe2018,VioTacTrash2019}.



A more common case for regrasping is the unstable grasp pose. In this case, the first step is to correct the grasp pose. The regrasp action is assumed as a translational action~\cite{Tac-Trans2018} and the tactile imprints after the regrasp action can be simulated and predicted. The proposed algorithm simplifies the regrasp actions only as translational action, which is not feasible when the object has shape of e.g., cylinder. Further, an action-conditional model with inputs of vision, tactile and grasp action is proposed to plan the regrasp~\cite{Gelsight2018}.

All the proposed regrasping methods are based on the tactile data collected by "touching" the static object, which has limitations for certain objects e.g., tool objects, where the single tactile data may not be sufficient for a robust prediction. Our proposed methods collected data by not only "touching" but also "lifting" the object.

To solve more specific problems in the domain of robotic grasping, a center-of-mass based grasp planning method is proposed by~\cite{Centerofmass}. Using a force/torque sensor and a 3D range sensor, the regrasp pipeline towards center of mass follows the measurement of torque  using a humanoid hand. Similarly, a regrasp policy by adding force in~\cite{dynamicCOM2016} is introduced to grasp objects with dynamic center of mass but only force instead of the grasp pose needs to be adjusted. In advance, we design a regrasp planner which predicts the grasp adaption based on slip detection using two-jaw gripper where objects are previously unknown.


\section{Problem Statement}
We attempt to solve the problem of planning a stable grasp pose with a parallel-jaw gripper for unknown objects laying with unknown mass distribution on the table inside the robot workspace, using a commercial RGB-D camera, tactile sensors and force/torque sensors. 

\begin{figure}[htp!]
\label{sec:method}
\centering
  \includegraphics[width=0.4\textwidth]{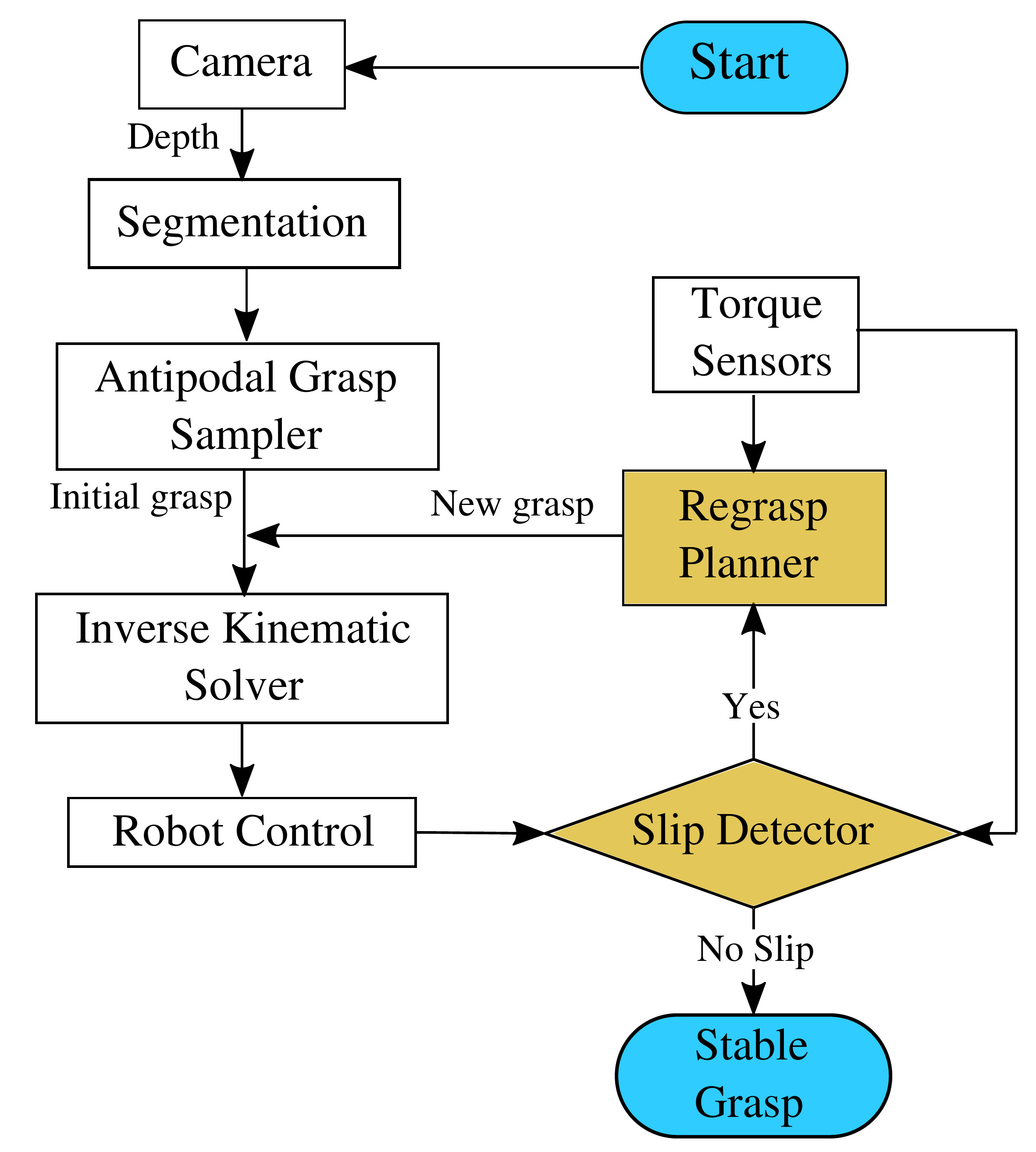}
\caption{The whole pipeline of the proposed method.}
\label{fig:flowchart}       
    \vspace{-6mm}
\end{figure}

\begin{figure*}[htp!]
\label{sec:method}
\centering
  \includegraphics[width=0.7\textwidth]{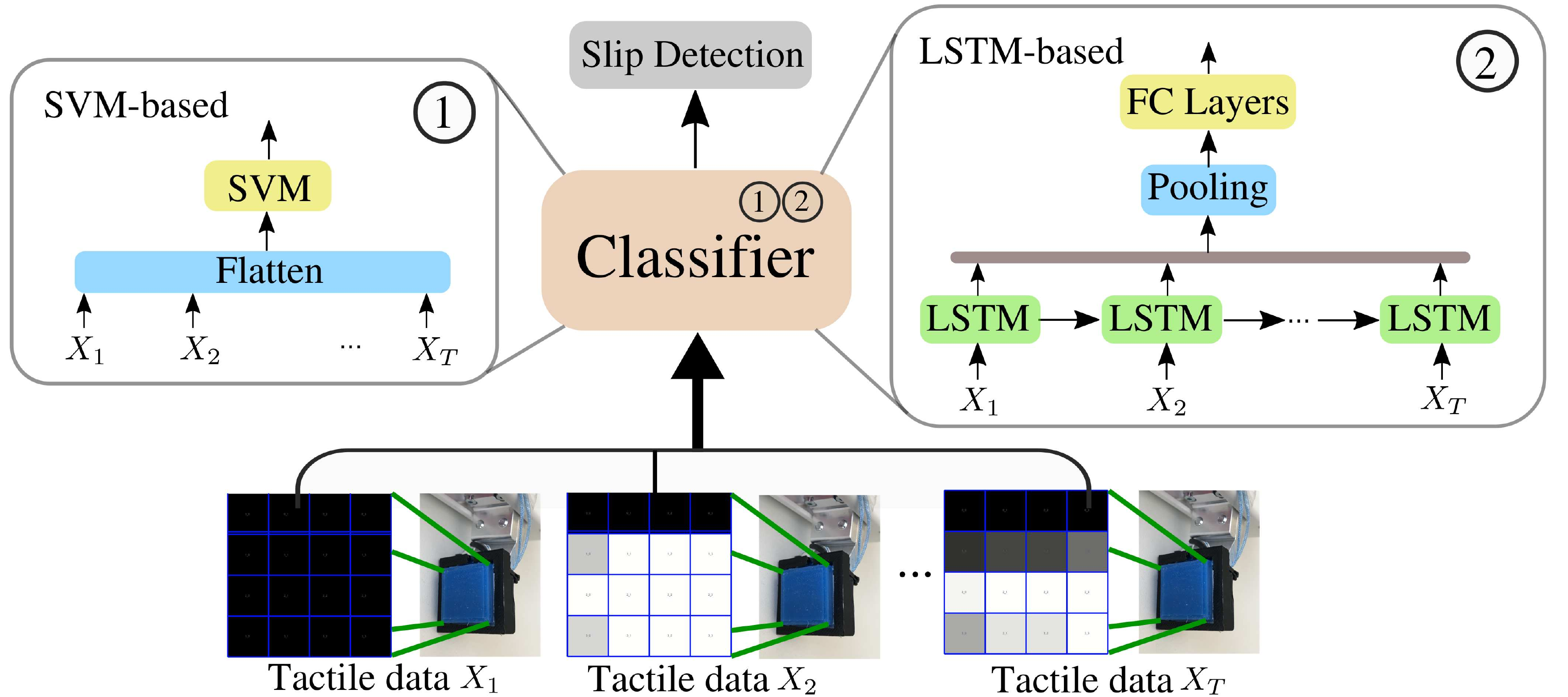}
\caption{The structure of the slip detection model. The input tactile data is visualized with a $4\times4$ GUI exhibiting the color intensity between black and white. Then the sequential tactile data ${X^n}_T$ is processed with 2 classifiers based on SVM and LSTM model to predict the slip. For SVM-based classifier, the sequential data is first flattened in a one-dimensional vector and then classified with SVM classifier. For LSTM-based classifier, the sequential data is processed with a LSTM layer of 75 memory cells. The extracted features are first fed into an average-pooling layer and then classified with two Fully Connected (FC) layers.}
\label{fig:lstm_model}       
    \vspace{-6mm}
\end{figure*}

\section{Methods}
\label{sec:methods}
The proposed grasping method is illustrated in~\Cref{fig:flowchart}.

\subsection{Antipodal Grasp Sampler}

The object is segmented out in a depth image using RANSAC plane segmentation method implemented in Point Cloud Library (PCL)~\cite{Ransac}. Afterwards with the object boundary, an antipodal grasp sampler is implemented to sample poses out of the object boundary and identify if the grasp pose is force-closure~\cite{Lizexiang}. The friction coefficient is assumed started with 0 and then it increases step by step to 1 with interval of 0.2 until a certain number of grasp samples are found. The depth value ``z'' of grasp pose is also sampled between the depth of table and the depth of the grasp center point. After choosing an initial grasp pose, the robot executes the pose using an inverse kinematic solver and robot joint motion control with the pipeline shown in~\Cref{fig:flowchart}.

\subsection{Slip Detector}

We attempt to learn a classifier $f(\cdot)$ named slip detector, given the $n_{th}$ tactile sensor data sequence $\{{X^n}_T: {x^n}_1,{x^n}_2,...{x^n}_T\}$ with a variable length $T$, which is collected during lifting the objects. The slip is classified with four classes $\{s_{no}:$``no slip'', $s_{cw}:$``clockwise rotational slip'', $s_{ccw}:$``counterclockwise rotational slip'', $s_{tra}:$``translational slip''$\}$, the first case is detected by determining if the object is still in contact with the tactile sensor and the other three cases are predicted by our classifier.
\begin{align}
s_t = f({X^n}_T), \label{slip_class}
\vspace{-6mm}
\end{align}
where $s_t \in [{s_{no},s_{cw},s_{ccw},s_{tra}}]$.

We compare the performance of two classification models to predict the slip.
\subsubsection{Support Vector Machine}

For the classification tasks, Support Vector Machine (SVM) finds a decision function efficiently with the maximal margin~\cite{SVM1992}. Given the training set of $T$ examples $x_i$ with labels $y_i$: ${(x_1,y_1),(x_2,y_2),...,(x_T,y_T)}$, where $y_k = 1$ means $x_k$ belongs to class A and $y_k = -1$ stands for class B. The algorithm finds the decision function $f(x)$~\cite{SVM1992}:
\vspace{-2mm}
\begin{align}
f(x) &= \sum_{k=1}^{T} w_k K(x_k,x)+b.  \label{decision_function} 
\end{align} 
In~\Cref{decision_function} $w_k$ is the weight, $K$ is the kernel function, $b$ is the bias and $x_k$ is a support vector. For 3-class prediction in our case, SVM draws decision function between each pair of classes so in total 3. Here we compare two kernels namely linear kernel and Radial Basis Function(RBF) kernel.


Other kernels such as polynomial and sigmoid kernels are less applied considering their computational reasons and more hyperparameters. Additionally, in~\cite{SVMSigmoid} the sigmoid kernel has very similar behavior like RBF kernel for certain parameters.

\subsubsection{LSTM}
Long short-term memory (LSTM) is one recurrent neural network which has shown great performance in processing sequential data, for example speech recognition~\cite{LSTMspeech,LSTMspeech2} or video recognition~\cite{LSTMvideo}.

The main problems of recurrent neural networks are the gradient vanish and gradient explosion problem~\cite{LSTM1997}. During the backpropagation, the gradient flow could vanish if it is frequently multiplied by a small value and it could also explode if often multiplied by a large value. The structure of LSTM model with three gate units could protect the access to the error flow, to enforce the error flow constant. Therefore we can keep the value of gradient in a proper range to avoid the problem of vanishing or exploding gradients~\cite{LSTM2000}.


Considering that the input tactile sequences may have variable lengths thus we resize all the sequences with the same length and then we employ a Masking layer from Tensorflow Keras before LSTM layers to mask those time steps for all downstream layers. The structure follows the pipeline shown in~\Cref{fig:lstm_model}.

\subsection{Learning a regrasp planner}
We attempt to utilize multiple sensor modules such as tactile sensors and torque sensors mounted on the robot arm to learn the regrasp. The unstable grasps in our experiment often lead to rotational slip because of too much torque generated from gravity. In other word, the grasp pose is far away from the center of mass. And the regrasp planner attempts to predict a stable grasp pose which is as close as possible to the center of mass.

\begin{figure}[!t]
\label{sec:regrasp}
\centering
  \includegraphics[width=0.3\textwidth]{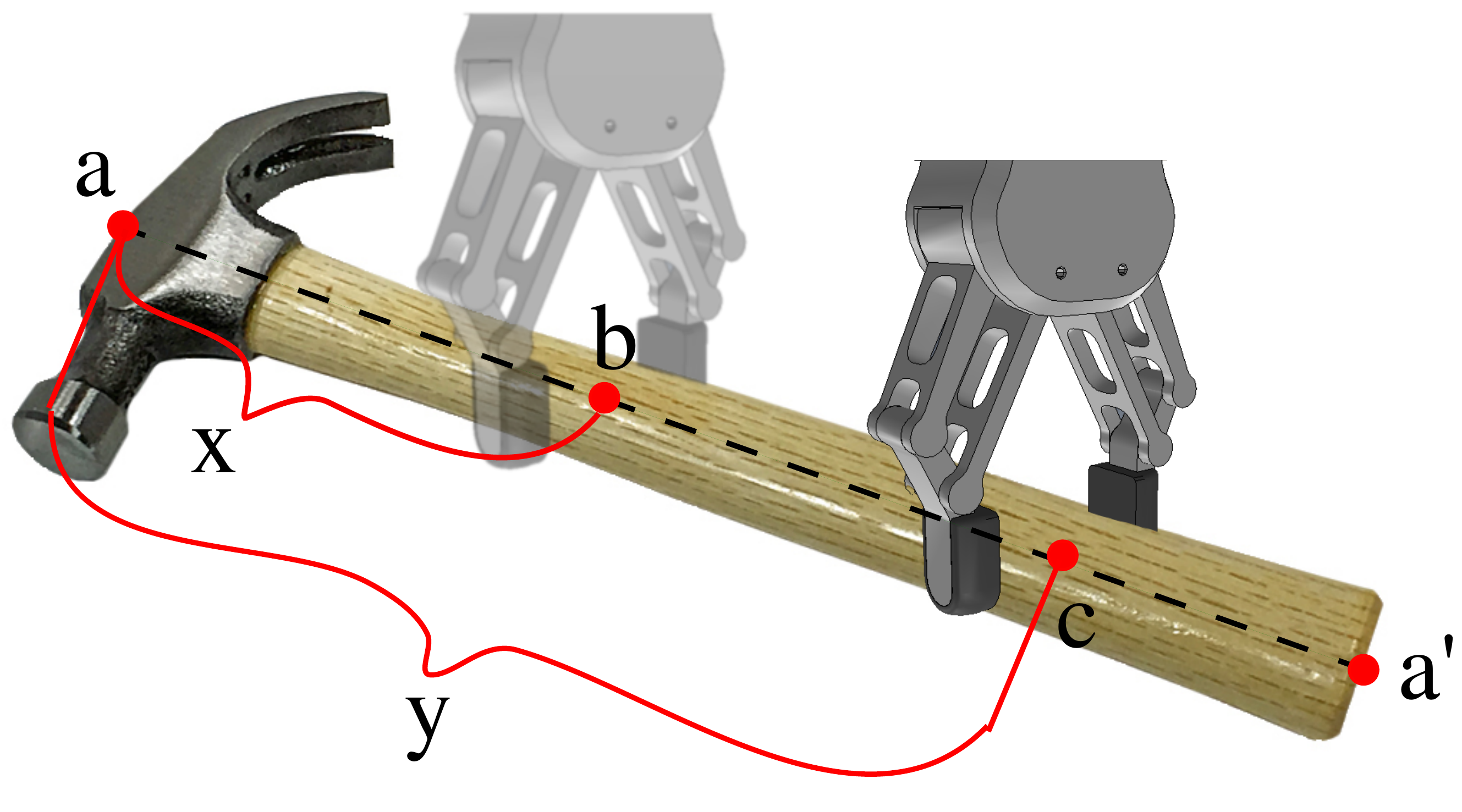}
\caption{Given the current grasp pose $c$, the object boundary point $a$, the regrasp pose is located with a variable called regrasp ratio $\mu$. Regarding the point $a$ as the reference point, the distance between $a$ and $c$ is $y$ and the distance between $a$ and $b$ is $x$. Thus the regrasp ratio $\mu = \frac{x}{y}$.}
\label{fig:regrasp_ratio}       
\vspace{-0.3cm}
\end{figure}

The regrasp action is formalized as an one dimensional adjustment. We use a variable $\mu$ to define the location of center of mass (i.e., the regrasp pose) based on the current grasp pose $c$ and the object boundary point $a$ visualized in~\Cref{fig:regrasp_ratio}. The object boundary points $a$ and $a'$ can be obtained by finding the points of intersection from the object boundary and the grasp normal which is depicted as dashed line in~\Cref{fig:regrasp_ratio}. One of the object boundary points will be chosen according to the rotational slip.

\begin{figure}[!t]
\label{sec:method}
\centering
  \includegraphics[width=0.5\textwidth]{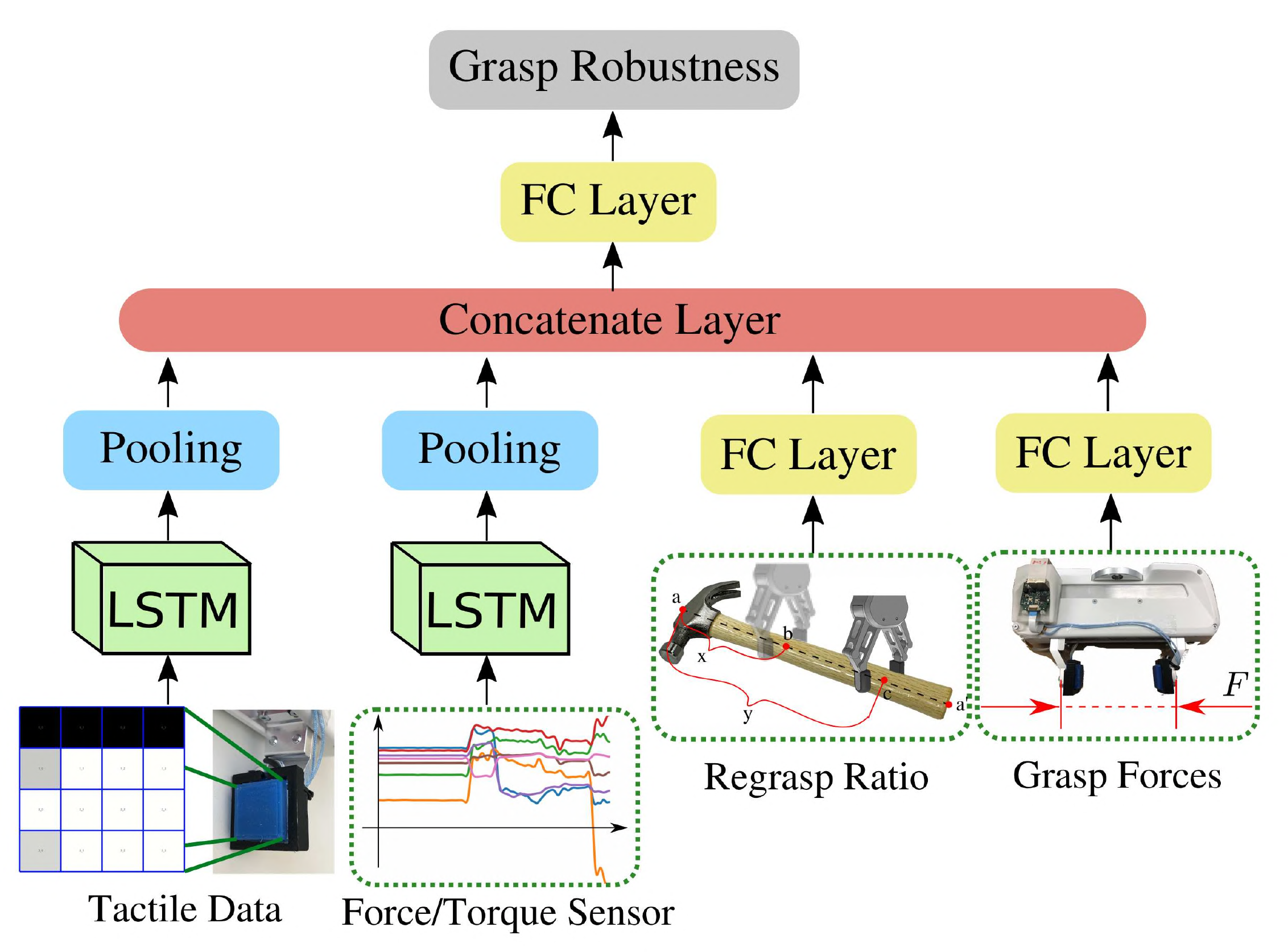}
\caption{The structure of the regrasp planner. The model takes the tactile sequential data, torque/force data, regrasp ratio and grasp force as inputs and it outputs the grasp robustness of the new grasp pose located according to regrasp ratio.}
\label{fig:ragrasp_model}       
    \vspace{-6mm}
\end{figure}

The whole regrasp planner model is illustrated in~\Cref{fig:ragrasp_model}. The first input is tactile sequential data with 32 features from 2 tactile sensors. 
The second input is also a sequential data of external forces and torques in 6-DOFs end effector frame. 
The third input is one scalar variable, the regrasp ratio described in~\Cref{fig:regrasp_ratio}. The fourth input is also a scalar variable, the adjustment of grasp force. 

We apply the LSTM model again to process the tactile sequential data and the torque/force data for regrasp planner. 
For the third and the forth scalar inputs, we simply add one dense layer to enlarge the output size. Afterwards the two LSTM layers and two dense layers get concatenated in the third dense layer along the axis of features. The whole features are classified with the following two dense layers and finally the grasp robustness is estimated.

With a learned regrasp planner, we sample the grasp ratio and choose one with the highest predicted grasp robustness. Given the object boundary point $a$ and the first grasp pose $c$, we can obtain the regrasp pose with equation :
\begin{align}
pose_b = pose_a + (pose_c - pose_a) * \mu.
\end{align}

\section{Experiments}
\label{sec:experiment}
\subsection{Experimental setup}



The robotic grasping experiment is conducted using a 7-DOFs Franka Panda robot arm, equipped with a Franka parallel-jaw gripper. The gripper is mounted with two tactile sensors on the fingertip.
The experiments are running on a laptop installed with Ubuntu 18.04 with a 4.1 GHz Intel Core i7-8750H 6-Core CPU and an NVIDIA GeForce GT 1060 graphic card.


The Franka parallel-jaw gripper in~\Cref{gripper_tactile} has one degree of freedom with an adjustable grasp  of $0.0cm$ to $8.0cm$. The force applied on gripper can be controlled in the range of $ \numrange[range-phrase = -]{20}{100} \si{\newton}$. The gripper during experiment is controlled to close until the required force is reached.

A commercial RGB-D camera Realsense D415 is mounted at a height of \SI{80}{cm} above the table and we capture depth images to sample the initial grasp poses.

\subsection{Tactile and Torque Sensors}
\begin{figure}[t!]
	\begin{center}
		\includegraphics[width=5 cm]{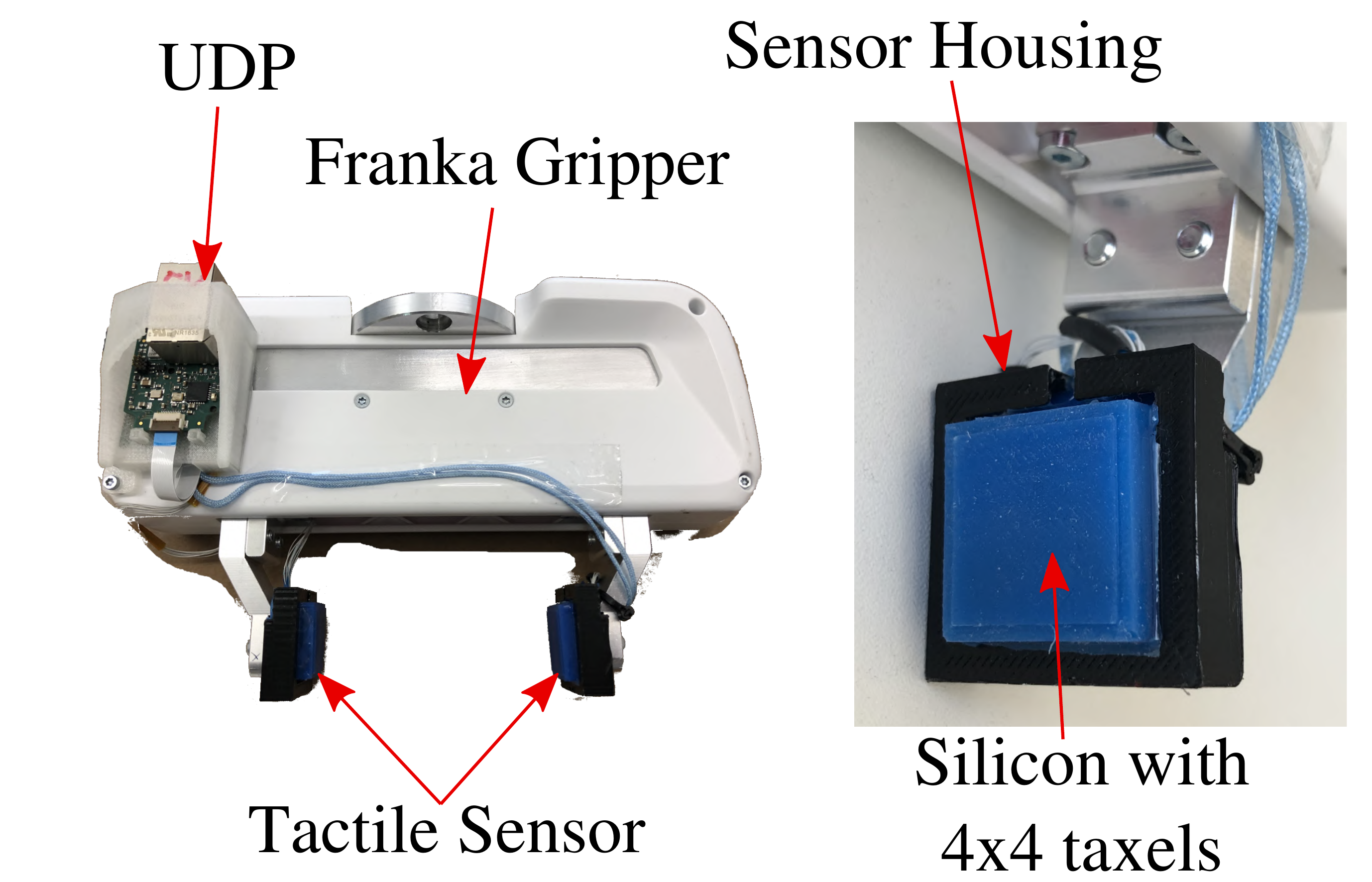}
		\caption{The Franka gripper and the tactile sensors employed for our experiment. }
		\label{gripper_tactile}
	\end{center}
\vspace{-6mm}
\end{figure}

The tactile sensor in~\Cref{gripper_tactile} provided by the Kinfinity UG\footnote{\url{http://kinfinity-solutions.com}}  is a 3D-printed low-resolution pressure tactile sensor with 4$\times$4 resolution, which provides the information about pressure distribution on the contact surface. 
It consists of a blue silicon cover and 4$\times$4 tactile taxels embedded inside, where each taxel has a spatial dimension of 4$\times$\SI{4}{\mm}. 


The sensor output is transmitted through User Datagram Protocol (UDP) protocol via Ethernet cable. An infinite impulse response (IIR) digital filter is designed to remove high-frequency noise in tactile data, which also are capable for real-time applications. 

The Franka Emika Panda platform uses 7-DOF joint torque sensors. The force/torque data in end effector frame is obtained by applying Jacobian matrix transformation.


The data is streamed at a frequency of \SI{1}{\kHz} and with filtering we collect the data at a frequency of \SI{50}{Hz}.


\subsection{Experimental objects}

The training dataset contains daily tool objects such as axes and hammers, box-like objects and also a modularized object, considering the limitations of the experimental hardware and software.
\begin{figure}[!htb]
\centering
   \subfigure[The experimental objects]{
    \includegraphics[width=0.35\columnwidth]{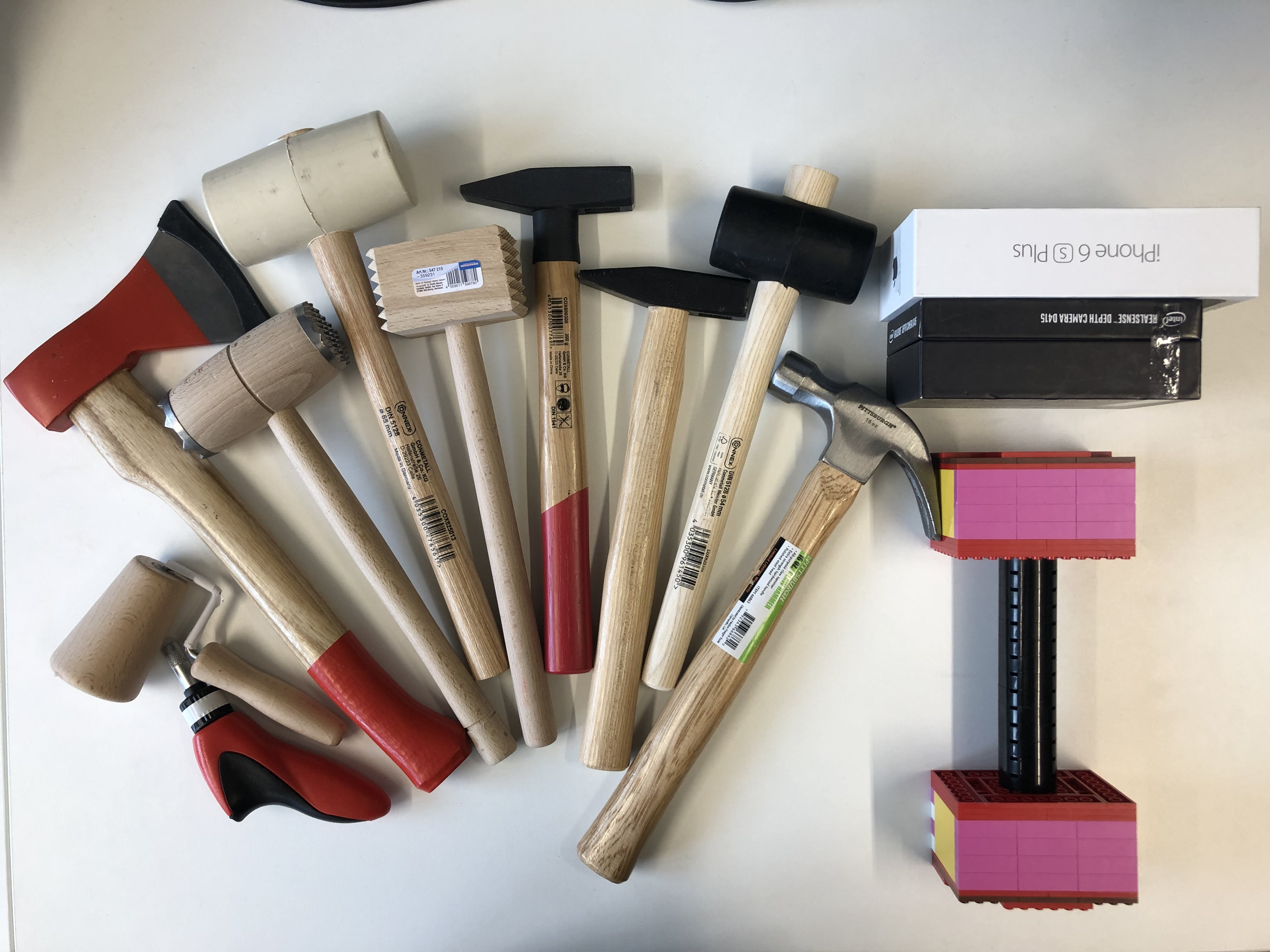}
    \label{fig:objects}
    }
     \subfigure[The tested objects.]{
    \includegraphics[width=0.35\columnwidth]{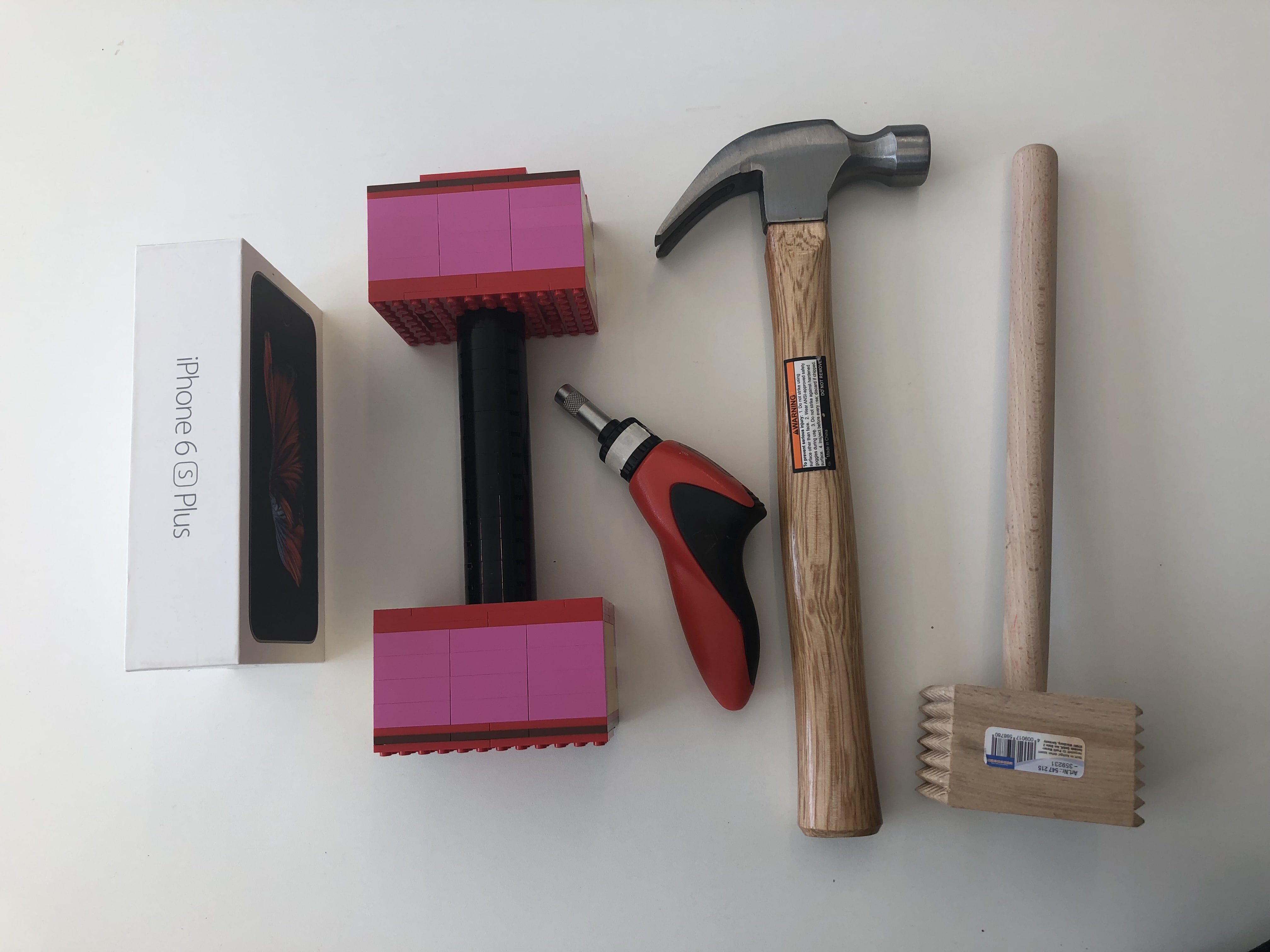}
    \label{fig:test_objects}
  }
     \subfigure[The ``open'' LEGO]{
    \includegraphics[width=0.35\columnwidth]{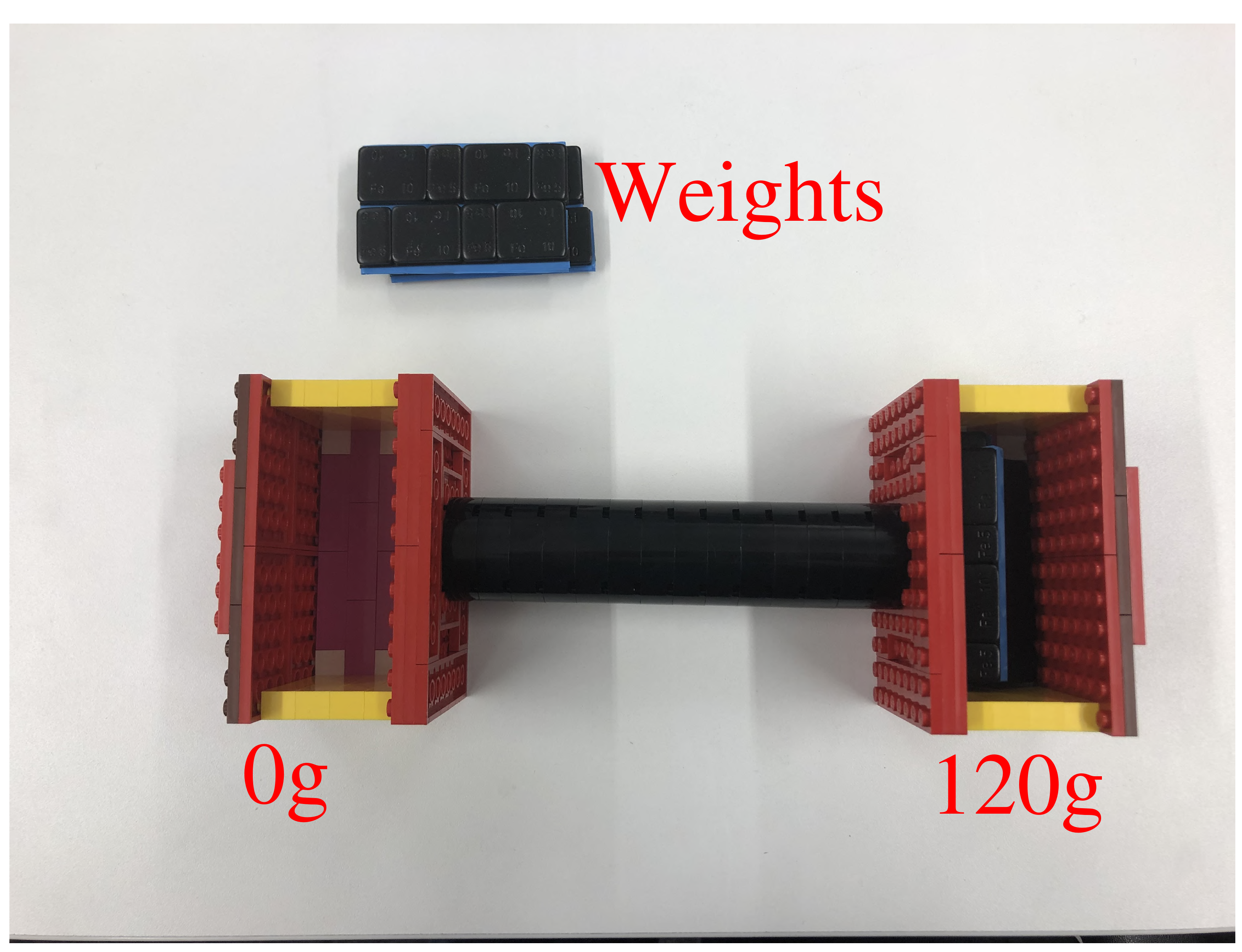}
\label{fig:lego}    
    }
     \subfigure[Th ``closed'' LEGO.]{
    \includegraphics[width=0.35\columnwidth]{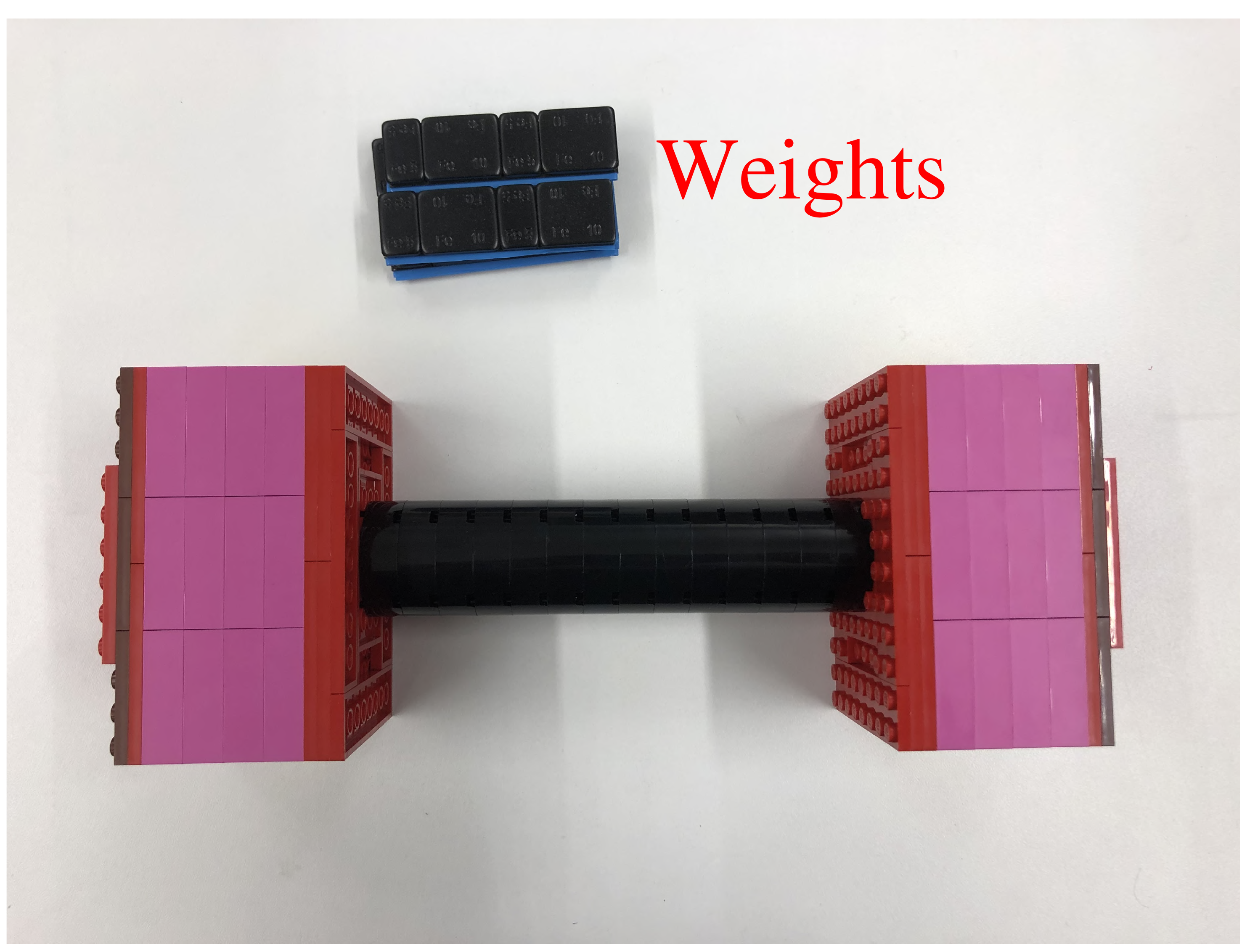}
  } 
  \caption{All objects used for our experiment. The objects are collected from tools of daily use, box-like objects and a modularized object made of LEGO. The modularized object allows us to load extra weights on the two sides. For example in (c) the weights on left and right side are \SI{0}{g} and \SI{120}{g}, or $(0,120）\si{g}$. The "closed" LEGO in (d) shows no differences in visual sensor but it may has totally different center of mass. Same works for box-like objects.}
\vspace{-6mm}
\end{figure}

With the limited number of experimental objects, we keep the material of objects same by adding extra weight to adjust the locations of center of mass as well as the object mass, such as the different configurations for LEGO model in~\Cref{fig:lego}.

In total, there are 6 configurations for training, namely $(0,0)$, $(120,0)$, $(120,120)$, $(240,0)$, $(240,120)\si{g}$. Additionally, a configuration of extra weights on left and right sides with $(60,300)\si{g}$ is used for the test phase. 

In summary, we collect totally 19 objects (originally 13 objects with different extra weights) including 12 objects of daily use and a modularized object made of LEGO model to generate 7 different configurations. We split those 19 objects into 14 training objects and 5 test objects.

\subsection{Data collection}
An automatic data collection process is designed to obtain training data. The robot will have a first trial to grasp the object 
and lift it to the constant height of \SI{10}{\cm}.
Then a random regrasp ratio $\mu$ is chosen to define the regrasp pose as illustrated in~\Cref{fig:regrasp_ratio}.


After two grasp trials the robot will change the pose of grasped object to another random pose and then release it on the table for the next experiment.

In the experiment there are only 14 samples labeled with ``translational slip'' out of 1039 samples so we will ignore this case in the result. Totally we collect 1039 grasps for the slip detection from 12 objects and 1\,347 regrasps.



For the regrasping, we label the data with ``1'' if the object is grasped successfully without slip, 
otherwise we label the data with ``0''.



\subsection{Training}

First, we preprocess the data with feature standardization method.Then we split the training dataset objectwise randomly into training set and validation set with a ratio of 5:1. 

For the SVM based slip detector, using 5-fold cross validation, the hyperparameter $C$ for linear SVM and RBF kernel SVM are chosen to be 1 and $1\mathrm{e}{3}$. The hyperparameter $\gamma$ in RBF kernel function. 

The LSTM based slip detector has a batch size of 16, 75 LSTM memory cells, a learning rate of $1\mathrm{e}{-3}$ and a hidden layer size of 50. To avoid overfitting we deploy dropout and recurrent dropout~\cite{DropoutLSTM} with rate of 0.2 for the LSTM model meanwhile a dropout layer with rate of 0.5 after the hidden FC layer.
We choose binary cross entropy method as loss function and Adam optimizer as the optimization method.
With similar hyperparameters, LSTM based regrasp planner uses mean squared error as loss function.

\section{Results}
\subsection{Slip Detection}
\begin{figure}[t!]
    \centering
    \includegraphics[width=6cm]{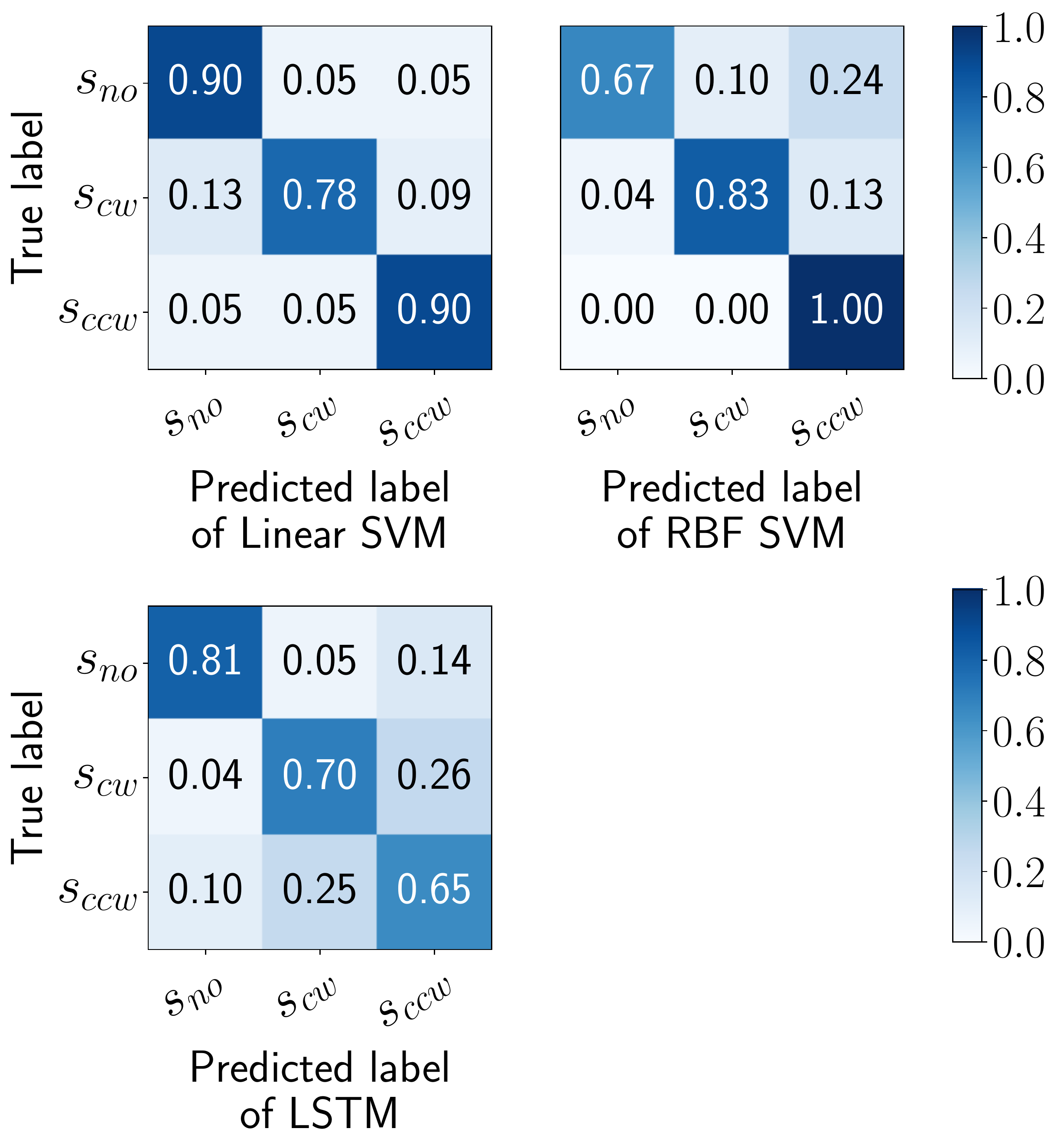}
    \caption{Comparison of three classifiers using confusion matrix. All prediction accuracy is normalized for each class. The linear SVM classifier has the best validation results.}%
    \label{confusion_matrix}%
\vspace{-6mm}
\end{figure}

The evaluation results are shown in~\Cref{confusion_matrix}.
Linear SVM outperforms other classifiers with an accuracy of 84.4\% (RBF kernel SVM with 82.8\% and LSTM classifier with 74.9\%). For the classification accuracy of each label, linear SVM and RBF kernel SVM both achieve best classification performance on the ``counterclockwise slip'' class each with an accuracy of 90.0\% and 100.0\% respectively. Meanwhile LSTM model relatively is better at detecting stable grasp of the class ``no slip'' with a 81\% accuracy.

Linear SVM performs slightly better than RBF kernel SVM. Since we flatten the tactile sequential data into an one-dimensional vector, the feature size increases from ``16'' to ``sample length $\times$ 16''. In case where the feature size is larger than the instances, RBF kernel could not outperform linear kernel because there is no need to project the data into a higher-dimensional feature space. 
Meanwhile, LSTM classifier does not perform as well as other classifiers, probably because of the limited dataset.

\begin{table*}[!h]
\centering
\vspace{4mm}
\caption{Regrasp planner test results}
\begin{tabular}{|c|c|c|c|c|c|c|c|c|}
\hline
\multirow{2}{*}{Algorithms} & \multicolumn{2}{|c|}{Random planner}  & \multicolumn{2}{|c|}{Dex-Net 4.0}  & 
\multicolumn{2}{|c|}{Simpler Regrasp Planner} & \multicolumn{2}{|c|}{Our Regrasp Planner}\\ 
\cline{2-9}
& success & success rate & success & success rate & success & success rate & success & success rate \\
\cline{2-9}
\hline
Hammer  & 2 & 10.0\%  & 0 & 0.00\%  & 7 & 35.0\%  & 16 & \textbf{80.0\%}  \\
\hline
Wooden hammer  & 16 & 80.0\%  & 17 & 85.0\%  & 19 & 95.0\%  & 20 & \textbf{100\%}  \\
\hline
iPhone box  & 5 & 25.0\%  & 6 & 30.0 \%  & 8 & 40.0\%  & 9 & \textbf{45.0\%} \\
\hline
LEGO model  & 11 & 55.0\%  & 12 & 60.0\%  & 16 & 80.0\%  & 18 & \textbf{90.0\%} \\
\hline
Screwdriver & 14  & 70.0\%  & 14 & 70.0\%  & 16 & 80.0\%  & 17 & \textbf{85.0\%} \\
\hline
Mean  & 9.6 & 48.0\%  & 9.8 & 49.0\%  & 13.2 & 66.0\%  & 16.0 &\textbf{80.0\%} \\
\hline
\end{tabular}
\label{online_regrasp}
\vspace{-5mm}
\end{table*}
\subsubsection{Torque Sensors for Slip Detection}



In~\Cref{torque_slip} we see that performance indeed drops significantly when force/torque data is involved. The possible reasons are on the one hand, torque sensors are known for inevitable noise and drift. On the other hand, the object can be grasped but still lying on the table because of a rotational slip thus the external forces from gravity are relative small. Meanwhile with torque input, LSTM model outperforms other SVM models with more than 10\% accuracy. It indicates that LSTM model can process multi-sensor input modules and extract features better than SVM models.

\begin{table}[!ht]
\centering
\caption{ K-fold(K=5) cross-validation accuracy of different slip detection models}
 \begin{tabular}{|c|c|c|c|}
\hline
\makecell{Slip Classifier} & \makecell{Linear SVM} & \makecell{RBF SVM} & \makecell{LSTM} \\
\hline
tactile only & \textbf{88.2}\% & 80.5\% & 82.1\% \\
\hline
torque only & 58.1\% & 61.8\% & \textbf{75.5}\% \\
\hline 
tactile + torque & 68.8\% & 68.1\% & \textbf{78.1}\%  \\
\hline
\end{tabular}
\label{torque_slip}
\vspace{-3mm}
\end{table}

\subsubsection{Robot Grasp Evaluation}
Afterwards we test our trained classifiers on the robot towards novel test objects with the following test objects listed in~\Cref{fig:test_objects}. 

\begin{table}[ht!]
\centering
\caption{ The $F_{\text{score}}$~\cite{scikit-learn} of slip detection from a linear SVM and a LSTM classifier}
 \begin{tabular}{|c|c|c|}
\hline
\makecell{Slip Classifier} & \makecell{Linear SVM} & \makecell{LSTM} \\
\hline
Hammer & 71.43\% & \textbf{82.98\%} \\
\hline
Wooden hammer & \textbf{90.50\%} &  75.00\% \\
\hline
iPhone box &  \textbf{82.35\%} & 56.70\% \\
\hline
LEGO model &  57.14\%  &  \textbf{62.07\%} \\
\hline
Screwdriver & \textbf{86.25\%} & 57.14\%\\
\hline  
Mean & \textbf{76.88\%} & 66.78\% \\
\hline
\end{tabular}
\label{online_test}
\vspace{-3mm}
\end{table}


The results from online test in~\Cref{online_test} are quite similar to the offline results. Linear SVM outperforms LSTM classifier with 10.10\% but they have different performances on the individual object. Linear SVM has better performance on most objects than LSTM classifier except hammer and LEGO model. The linear SVM classifier with best performance will be applied for our regrasp planner.

\subsection{Regrasp Planner}

We evaluate the regrasp planner with different input modules in~\Cref{torque_regrasp}. The result shows a better performance of inputs with multi-sensor modules.

\begin{table}[h!]
\centering
\caption{ Different input modules for regrasp planner}
 \begin{tabular}{|c|c|c|c|}
\hline
Input modules & Tactile only &Torque only & Tactile + torque \\
\hline
Accuracy & 66.8 \% & 63.3 \% &\textbf{75.2\%} \\
\hline
\end{tabular}
\label{torque_regrasp}
\vspace{-3mm}
\end{table}

We use grasp success rate to compare different policies. The definition of a successful grasp is that only if the object is lifted stably without any slip, determined by a human expert. With the limited grasp force and relative heavy experimental objects chosen by us, the grasp is likely to be unstable if the grasp pose is not close enough to the center of mass. 

\begin{itemize}
	\item Random grasp planner: The grasp pose is randomly chosen from the antipodal grasp sampler. 
	\item Dex-Net 4.0~\cite{dex4.0}: A grasp planning model trained on Dex-Nex 4.0 dataset using a parallel-jaw gripper.
	\item Our slip detector + a simple regrasp policy: A simple regrasp policy with a fixed regrasp ratio of 0.5.	
	\item Our slip detector + our regrasp planner: Our learned regrasp planner using multi-sensor modules to plan a stable grasp based on slip detector.
\end{itemize}
	The first two policies are open-loop and the rest policies with regrasp planner are closed-loop with the feedback of the
    slip detection.

Each policy is evaluated with two grasp trials and each object is grasped with 20 times. For the regrasp planner, both the false prediction from slip detector and regrasp planner will lead to a failure.

In~\Cref{online_regrasp}, the result implies that the closed-loop policies outperform the open-loop policies. The feedback from slip detection in our case can help to regrasp and thus improve the grasp robustness.

For the open-loop policies, Dex-Net 4.0 does not show much better performance than a random planner. Because the Dex-Net is trained on a dataset where all objects are assumed with even mass distribution. In our case, many objects e.g., hammer, LEGO model and screwdriver have uneven mass distributions. Second, all experimental objects are relative heavier with our experimental setup, which makes it critical for a grasp pose to be close to the center of mass. However, Dex-Net 4.0 predicts most of the poses close to the pixel-wise center in the depth image instead.

For the closed-loop policies, the learning based regrasp planner with predicted regrasp ratio outperforms the Dex-Net 4.0 with 31\% and the simpler regrasp planner with 14\% success rate. 



\section{Conclusion \& Future Work}
\label{sec:conculsion}
We present a novel learning based approach using multi-sensor modules to predict stable grasp poses of unseen objects, based on the slip detection. The proposed approach consists of two parts, a slip detector and a regrasp planner. Both models are learned from real-world experiments with ground-truth label. An initial antipodal grasp pose is chosen to be executed. Then the slip is detected by a slip detector during lifting the object. Afterwards a regrasp planner predicts a new stable grasp pose by predicting the location of center of mass based on the feedback of the slip detector and also other sensor modules. We demonstrate that our learning based slip detector and regrasp planner effectively generalizes learned knowledge to detect slip and regrasp while grasping novel objects. 
We believe our proposed algorithm can potentially be an Add-on algorithm for general grasp planners used for applications such as pick and place.

The proposed algorithm can only correct the grasp pose in one direction where the rotational slip happens. This functionality has limitations for objects with complex shapes, thus the dataset and the number of objects are limited. In the future, we will attempt to collect more data either by a better automatic self-supervised data collection process or by generating synthetic data from simulation. Also a more general regrasp algorithm will be studied for objects with complex geometries, to detect slip and plan regrasp.
%
%

\section*{ACKNOWLEDGMENT}
This research has received funding from the European Union’s Horizon 2020 research and innovation programme under the Marie Sklodowska-Curie grant agreement No 691154 STEP2DYNA and No 778602 ULTRACEPT.
We thank the products offered from Maximilian Maier and Dr. Maxime Chalon. We thank Yunlei Shi, Kaixin Bai, Lei Zhang, Manuel Brucker and Karan Sharma for excellent advises and proofreading.


\bibliographystyle{IEEEtran}
{\scriptsize
\vspace{0.01 cm}
\bibliography{MyCollection}
}

\end{document}